# STR-GQN: Scene Representation and Rendering for Unknown Cameras Based on Spatial Transformation Routing

Wen-Cheng Chen
National Cheng Kung University
jerrywiston@mislab.csie.ncku.edu.tw

Min-Chun Hu
National Tsing Hua University
anitahu@cs.nthu.edu.tw

Chu-Song Chen
National Taiwan University
chusong@csie.ntu.edu.tw

## Abstract

*Geometry-aware modules are widely applied in recent deep learning architectures for scene representation and rendering. However, these modules require intrinsic camera information that might not be obtained accurately. In this paper, we propose a Spatial Transformation Routing (STR) mechanism to model the spatial properties without applying any geometric prior. The STR mechanism treats the spatial transformation as the message passing process, and the relation between the view poses and the routing weights is modeled by an end-to-end trainable neural network. Besides, an Occupancy Concept Mapping (OCM) framework is proposed to provide explainable rationals for scene-fusion processes. We conducted experiments on several datasets and show that the proposed STR mechanism improves the performance of the Generative Query Network (GQN). The visualization results reveal that the routing process can pass the observed information from one location of some view to the associated location in the other view, which demonstrates the advantage of the proposed model in terms of spatial cognition.*

## 1. Introduction

Understanding the structure of 3D scenes from the 2D observations is a fundamental topic in the field of computer vision. With the progress of the geometry-based model, researchers have developed several techniques to recover the 3D geometry from 2D views via optimization [10, 13] and machine learning [21, 24, 25, 20, 4, 27, 26, 5]. Different from reconstructing the explicit 3D geometry, Generative Query Network (GQN) [3] constructs the implicit scene representation and achieves novel view rendering merely based on the observed images and the pose information.

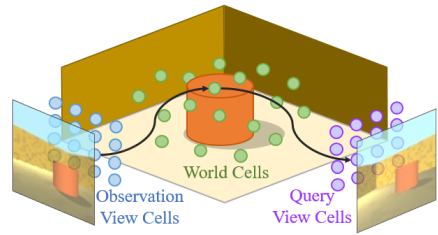

Figure 1. The basic concept of the proposed model, which treats the spatial transformation as a message passing process.

Unfortunately, GQN does not consider the spatial property and only has a weak generalization ability. For example, GQN fails to generalize the knowledge learned from the scene containing two objects to the scene containing four objects. Based on the geometric optics and camera models, some recent works combine the geometry-aware operations with GQN so that the model can be applied to more complex scenes. For instance, E-GQN [19] utilizes epipolar line to search the features in the observations. GRNN [22] applies camera projection and unprojection to pass the information between 2D views and a voxel-based feature memory.

However, geometry-aware operations rely on accurate camera intrinsic parameters which require additional calibration process and can only be used in a fixed and simple imaging situation. To design a general model which can be flexibly applied to different visual sensors with different imaging situations, an interesting question thus arises: can the model learn the spatial transformation property of the 3D space through observations without applying explicit geometry priors such as camera projection matrix, field of view, and the distortion coefficients? Aiming at this goal, we design a new architecture based on GQN to separate the process of spatial transformation and feature extraction, so as to learn the content-independent concept of spatial cognition. Following this idea, we propose a **Spatial Trans-** formation Routing (STR)** mechanism, which treats the transformation between world space and view space as a message-passing process. As shown in Fig. 1, the observed features are first passed to several locations in the world space. After fusing the features of different observations in the world space, the features are then passed to the view space of the query pose. The relations between the view poses and the routes of message passing are modeled by a neural network, which is end-to-end trainable.

Besides, after investigating the scene fusion operation and the associated meanings of the extracted features, we develop **Occupancy Concept Mapping (OCM)** to fuse the features of different observations with a probability model and constrain the scale of scene representation. Leveraging STR mechanism and OCM, we introduce **Spatial Transformation Routing Generative Query Network (STR-GQN)** (illustrated in Fig. 2) to achieve scene representation and rendering. The evaluation results on several benchmarks reveal that the proposed STR-GQN improves the performance of the baseline GQN model and achieves higher generalization ability. Moreover, we visualize the message passing results of Gaussian signals to disclose and explain how our model achieves spatial transformation.

## 2. Related Works

### 2.1. 3D Reconstruction

Constructing 3D models from multi-view images is a classic problem that has been widely explored in the field of computer vision. Structure from motion (SfM) [13] or simultaneous localization and mapping (SLAM) [10] methods optimize the re-projection error to estimate the ego-motion and build the 3D point cloud of the scene. Recent deep learning researches have taken the geometry properties into account to keep the spatial transformation consistency in 3D space for the tasks of 3D reconstruction [21, 24, 25, 20] and depth estimation [4, 27, 26]. For example, some works [24, 25, 20] generate the voxel-based structure and utilize differentiable camera projection to predict the 2D view and compute the re-projection loss for preserving the 3D-2D consistency. Other works [4, 27, 26] generate the warping map via the depth and ego-motion, and then utilize the differentiable warping operation to reconstruct the images of different views.

Aiming at structure reconstruction with uncalibrated cameras, Gordon et al. [5] learns the intrinsic parameters and estimates the depth from the video. However, their work requires successive frames with large overlapping areas as the input, while ours can be applied on randomly selected frames. Moreover, instead of learning the intrinsic parameters of the camera model, we learn the routing path for message passing, which is a more general mechanism and can be applied to different kinds of visual sensors.

### 2.2. View Synthesis

View synthesis aims to generate novel views of a scene given one or more images. Based on DNNs, Tatarchenko et al. [17] directly generate the novel view image from a single view without applying ground truth depth or 3D geometry information. The later works [28, 12] generate warping flow instead of the RGB image to achieve higher perceptual property. Some recent works [1, 23] further consider the geometric structure of the scene by projecting the generated depth map to construct the warping flow. As for the multi-image case, Sun et al. [16] predict the flow and image of a novel view for each input view, and aggregate the information to generate the final results. Choi et al. [2] generate the predicted depth probability volume of each view and fuse them into the depth probability volume of the target view. We will introduce other multi-image view synthesis methods which are based on neural scene representation and rendering in the following subsection.

### 2.3. Neural Scene Representation and Rendering

Neural scene representation and rendering models learn implicit representations of the scenes by training an end-to-end neural network to predict the images of novel viewpoints. Related works can be roughly divided into two categories according to the application scenarios. The first scenario aims to model the variation of a scene (e.g. the objects with different colors or be placed with different poses) with appearance distribution based on few observations. For example, GQN [3] concatenates the individual view-pose vectors with the image feature to extract the scene representation. By summing the scene representation of different poses, the global scene representation is then constructed and taken as the condition of the generative model to render the image for a novel view. Based on GQN, E-GQN [19] applies an Epipolar Cross-Attention mechanism, which leverages the epipolar constrain to perform non-local attention that can help the model to render more complex scenes. Another similar work is GRNN [22], which utilizes differentiable camera projection and unprojection operations to pass the features between 2D views and a voxel-based feature memory. The second scenario aims to record the details of a fixed scene and construct the representation by rich observations. For example, some works [14, 8, 11] construct a volumetric scene representation by ray tracing and project the volumetric feature to construct the feature map of the query view. SRN [15] and Nerf [9] adopt a quite different strategy, which directly learns the mapping from query location to light information and stores the scene information at the weighting of the network. This strategy can achieve infinite spatial resolution for the scene representation and result in high-quality scene rendering performance.

Our work mainly targets on the first scenario, and the most relevant work is GRNN [22], which applies the 2D-3D

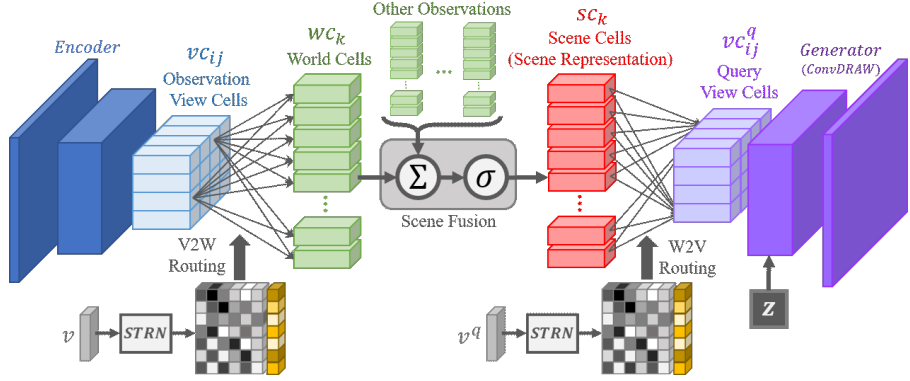

Figure 2. The architecture of the proposed STR-GQN model.

consistency to model the spatial property. However, GRNN adopts fixed camera projection/unprojection operation for the message passing process, while our model learns the message passing process directly from observations.

## 3. Method

The goal of our work is to achieve neural representation and rendering only based on the images and the corresponding poses without knowing the camera intrinsic parameters, which is the same as the problem setting in GQN [3]. We propose a possible architecture to learn the property of spatial transformation from observations. As shown in Fig. 1, the cognition of 3D scene can be represented by "world cells" that store the information of several specific locations in the world space. Once we receive the information from visual sense, the information will be stored in "observation view cells" arranged in a two-dimensional space. The information of observation view cells is then passed to world cells to construct the scene representation. Given a query pose, the information of world cells will be passed to "query view cells" arranged in a two-dimensional space to generate the query image.

In the beginning, the model does not know how the information is passed between different types of cells, and each routing edge has the same weighting. After exploring the environment based on different poses, "Coincidence" can be found in the observations. That is, similar information might appear at different positions (i.e. pixels) in different view spaces. Once the coincidence occurs, all weights on the routing paths connecting the observation view cell and the corresponding query view cell will increase. After modifying the weights based on several observations, the routing paths for message passing can more precisely reflect the property of spatial transformation.

To model the above process of spatial transformation learning, we propose **Spatial Transformation Routing Generative Query Network** as illustrated in Fig. 2. We utilize multiple convolution layers as the encoder to extract the features of observation view cells, denoted as $vc_{ij}$. The world cells are denoted as $wc_k$, and the scene cells which fuse the information of the world cells based on different observations are denoted as $sc_k$. Each cell is represented as a $c$-dimensional vector. The routing weights between view cells and world cells are generated by the spatial transformation routing network (STRN), which takes the camera pose of the observation view $v$ or the query view $v^q$ as the input. A convolutional DRAW model [6] is employed as the generator to reconstruct the image based on the query view cells, which are denoted as $vc_{ij}^q$. The details of the spatial transformation routing and scene fusion mechanism will be described in the rest of this section.

### 3.1. Message Routing for Spatial Transformation

Let $R$ be the relation matrix for the view cells and world cells. Each element $R_{ij,k}$ represents the relation between the view cell at the position $(i,j)$ and the $k$th world cell, i.e. relation between $vc_{ij}$ and $wc_k$. The view-to-world routing process is formulated as the weighted sum of the view cells based on the probability distribution $p_k^{dist}(i,j)$ defined by the normalized exponential of the relation $R_{ij,k}$:

$$p_k^{dist}(i,j) = \frac{\exp(R_{ij,k})}{\sum_{i',j'} \exp(R_{i'j',k})},$$
$$wc_k = \sum_{i,j} p_k^{dist}(i,j) vc_{ij}. \quad (1)$$

The above formulations have the drawback that the signals of all world cells have the same scale due to the normalization of the probability distribution. However, the scale of the signal might not be the same for different world cells. As illustrated in Fig. 3, the back-projection area of a given pixel increases with the viewing depth. Suppose the integral of signal magnitude on the covered area of a pixel would be the same for different viewing depths and the world cells are fixed size in the world space, the magnitude of the signal passing through a world cell will decay with the increase of the viewing depth. Furthermore, the out-of-view world cells should not receive any signal. To model the above phenomenons, we adopt an additional activation term $p_k^{act}$ named "frustum activation" to represent the scale of the signal for each world cell. The formulation of view-to-world routing process then can be formulated as:

$$wc_k = p_k^{act} \sum_{i,j} p_k^{dist}(i,j) vc_{ij}. \quad (2)$$

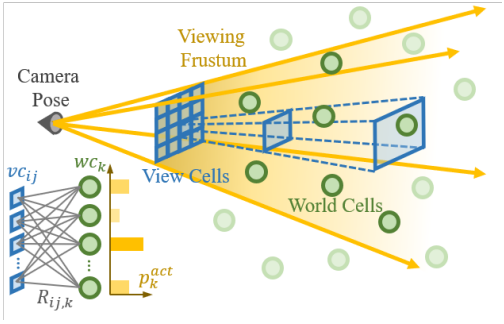

Figure 3. Concept illustration of the relation matrix $R_{ij,k}$ and frustum activation $p_k^{act}$.

The world-to-view routing process takes the scene cells masked by the frustum activation as the input to construct the query view cells. Note that the information of each scene cell is fused by the corresponding world cell of different observations, thus the scene cells and the world cells share the same space. Similar to the view-to-world routing process, the world-to-view routing process is formulated by the weighted sum of the masked scene cells based on the probability distribution $p_{ij}^{dist}(k)$ defined by the normalized exponential of the relation $R_{ij,k}$:

$$p_{ij}^{dist}(k) = \frac{\exp(R_{ij,k})}{\sum_{k'} \exp(R_{ij,k'})},$$
$$vc_{ij}^q = \sum_k p_{ij}^{dist}(k) p_k^{act} sc_k. \quad (3)$$

### 3.2. Spatial Transformation Routing Network

The relation matrix $R$ and the frustum activation $p_k^{act}$ change according to different view poses. The mapping between the view pose and the routing weights is modeled as a neural network named Spatial Transformation Routing Network (STRN). The proposed STRN aims to project the 3D camera space location of the world cells and the 2D view space location of the view cells to a common "spatial embedding space". The relation between a world cell and a view cell at a given view pose is computed by the similarity of their spatial embedding vectors. As illustrated in Fig. 4, the proposed STRN consists of three sub-networks:

**World-to-Camera Location Network (net$^{w2c}$).** This network takes the view pose $v$ as the input to extract the location codes of the world cells in the camera space. Note that each location code $c_k^{wc}$ is related but not equal to its location in the real world. The output of this network is a $3*K$ dimensional vector ($K$ is the number of world cells), which is then reshaped to $K$ location codes with 3 dimensions.

**World Cell Spatial Embedding Network (net$^{wce}$).** For each world cell, $net^{wce}$ utilizes its location code in camera space $c_k^{wc}$ to predict the corresponding frustum activation $p_k^{act}$ and the spatial embedding vector $e_k^{wc}$.

**View Cell Spatial Embedding Network (net$^{vce}$).** For each view cell, the location code in view space $e_{ij}^{vc}$ is constructed by the corresponding 2D-coordinate normalized into $[-1, +1]$. $net^{vce}$ takes the location code of each view cell to generate the spatial embedding vector $e_{ij}^{vc}$.

The relation matrix $R$ is then constructed by the inner product of the spatial embedding vectors of the view cells and the ones of the world cells. The forward process of STRN is written as follows:

$$c_k^{wc} = net^{w2c}(v),$$
$$e_k^{wc}, p_k^{act} = net^{wce}(c_k^{wc}),$$
$$e_{ij}^{wc} = net^{vce}(c_{ij}^{vc}),$$
$$R_{ij,k} = (e_k^{wc})^T (e_{ij}^{vc}). \quad (4)$$

### 3.3. Occupancy Concept Mapping for Scene Fusion

By fusing the features from different observations, the scene representation imply more specific attributes of each object such as shape and position. GQN adopts simple addition operations for scene fusing, which results in the inconsistent scale of scene representation given different numbers of observations. GRNN [22] adopts recurrent neural networks (RNNs) to update the scene representation, but the computation of RNNs can not be paralleled. Moreover, no explicit mathematical explanations are provided to support the above-mentioned fusion mechanisms. Inspired by the occupancy grid mapping algorithm in the robotic field [18], we propose an **Occupancy Concept Mapping (OCM)** framework, which not only gives a mathematical explanation for the scene fusing operation but preserves the scale of the scene representation.

Let $o_{k,c}$ denote the random variable that a concept exists at the 3D location of the $k$th world cell (i.e. $o_{k,c} = 1$) or not (i.e. $o_{k,c} = 0$), and $o_{ij,c}$ denote the random variable that a concept $c$ exists at the 2D location $(i,j)$ of a view cell. The goal of OCM is to estimate the scene cell $sc_{k,c}$, which represents the posterior of the existing probability for a concept $c$ at the location of the $k$th world cell given the observed images $x^{1:N}$:

$$sc_{k,c} = p(o_{k,c} = 1 | x^{1:N}) \quad (5)$$

The log-odds of the concept existing probability in world

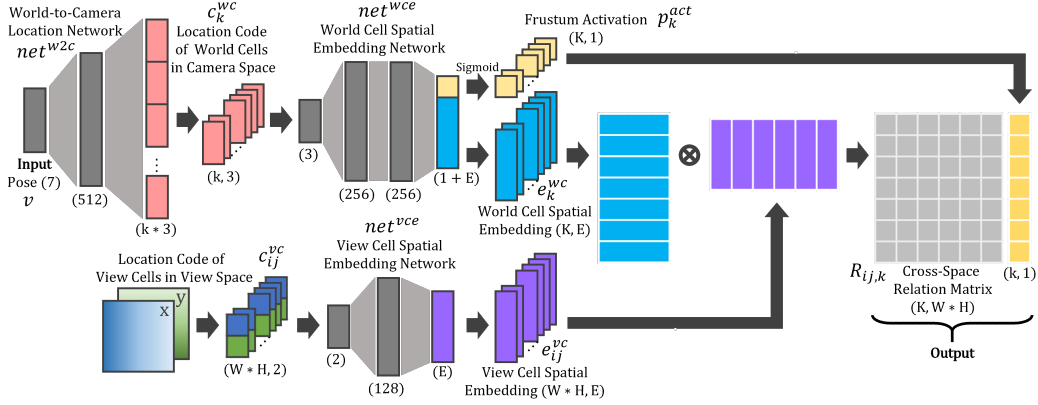

Figure 4. The architecture of spatial transformation routing network (STRN), in which $W$ and $H$ denote the width and height of the view cells, $K$ denotes the number of world cells, and $E$ denotes the dimension of the spatial embedding.

space is defined by:

$$\log Odd(o_{k,c}) = \log \frac{p(o_{k,c}=1)}{p(o_{k,c}=0)}. \qquad (6)$$

Let $wc_{k,c}$ denote the value of the $c$th channel for the $k$th world cell, which is assumed as the log of likelihood ratio of a concept:

$$wc_{k,c} = \log \frac{p(x|o_{k,c}=1)}{p(x|o_{k,c}=0)}. \qquad (7)$$

The scene cell is calculated by the posterior of log-odds:

$$sc_{k,c} = \sigma(\log Odd(o_{k,c}|x^{1:N})) = \sigma(\sum_{n=1}^{N} wc_{k,c}^n). \qquad (8)$$

Based on Eq. 8, the overall procedure of scene fusion can be formulated as the summation of world cells for different observations followed by a sigmoid activation function, which transforms the log-odds form to the probability form and preserves the scale of scene representation. The details of the computation for posterior and the analysis of the routing process in the probabilistic perspective are described in the supplementary materials.

### 3.4. Loss Function

The loss function is similar to the one used in GQN. Let $x^q$ and $x^q_{rec}$ denote the ground truth and the reconstructed query image respectively. $\mathcal{N}(\mu_l^g, \sigma_l^g)$ and $\mathcal{N}(\mu_l^e, \sigma_l^e)$ denote the prior and posterior of normal distribution latent code at the $l$th layer in the convolutional DRAW model respectively. The loss function $L$ consists of two parts, the reconstruction likelihood $L_{rec}$ and the regularization term $L_{reg}$:

$$\begin{aligned} L &= L_{rec} + \gamma L_{reg} \\ &= MSE(x^q_{rec}, x^q) + \gamma \sum_l D_{KL}(\mathcal{N}(\mu_l^e, \sigma_l^e)||\mathcal{N}(\mu_l^g, \sigma_l^g)), \end{aligned} \qquad (9)$$

where $\gamma$ denotes the weight of regularization term and is set to 0.001 in this work.

## 4. Experiments

### 4.1. Experimental Setup

**Datasets.** Five datasets used in GQN [3] and E-GQN [19] are adopted to evaluate our method, including Rooms-Ring-Camera (RRC), Shepard-Metzler-7-Parts (SM7), Rooms-Free-Camera (RFC), Rooms-Random-objects (RRO), and Disco-Humanoid (DISCO). The image size of data in RRC and SM7 are $64 \times 64$ pixels, and the image size of data in RFC, RRO, and DISCO are $128 \times 128$ pixels. Furthermore, we adopt the ShapeNet dataset used in GRNN [22] for evaluating the generalization ability of the proposed model. The image size in ShpaeNet dataset is $128 \times 128$ pixels. The training and testing sets are the images of scenes containing two objects. There is an additional evaluation set for the scenes containing four objects.

**Training Setting.** We randomly select one to five observation images in the training process and always adopt three observation images for evaluation. We train the network by using Adam optimizer with a learning rate of $5 \times 10^{-5}$. For each dataset, we adopt 0.1M samples for training with a batch size of 32 and totally 1.6M training steps (about 300 epochs). The experiments are conducted on a PC with Intel Xeon W-2125 CPU and NVIDIA RTX TITAN GPU.

**Model Parameters.** The width and height for view cells are 1/4 of that for the input image. The total number of world cells is set as 1024 and 2048 for images with $64 \times 64$ pixels and $128 \times 128$ pixels, respectively. The number of channels of the cells is set to 128, the dimension of the spatial embedding in STRN is set to 32, and the draw step of convolutional DRAW is set to 6.

**Comparison Models.** We adopt GQN [3] as the baseline model to evaluate whether the proposed STR mechanism improves the performance of the rendering results. Furthermore, we compare the proposed model with E-GQN [19] and GRNN [22], which adopt camera calibration parameters and geometry-aware operations to model the spatial property. We expect the proposed STR-GQN can outperform GQN and as good as the E-GQN or GRNN.

## 4.2. View Predictions

We first evaluated the reconstruction error of view prediction. Tab. 1 shows the root mean square error and mean absolute error of GQN [3], E-GQN [19] and our proposed STR-GQN. Another experiment was conducted to compare the proposed model with GRNN [22] because GRNN adopts a different setting that removes the stochastic unit and applies the cross-entropy pixel matching loss. We trained the proposed model and GQN model under the same setting of GRNN and evaluate the cross-entropy error as shown in Tab. 2. The experimental results show that the performance of STR-GQN outperforms the baseline GQN model in almost every dataset, which proves the effectiveness of our method in terms of modeling the spatial transformation. The proposed model with STR mechanism can achieve similar or even better performance compared to GRNN, which also adopts the concept of message passing between 2D view and 3D space but models the process via camera projection/unprojection. The performance of STR-GQN is marginally lower than E-GQN. The reason might be that E-GQN directly searches the feature on the observation images, which does not be affected by spatial resolution. In contrast, the performance of the proposed STR-GQN is limited to the number of world cells. Fig. 5 demonstrates the rendering results of the proposed method and GQN on each dataset. The proposed STR-GQN achieves better rendering quality than GQN and preserves the spatial rationality of each object in the scenes. However, the results on RRO and DISCO datasets reveal the limitation of our work. Although the location of each object is roughly correct, STR-GQN fails to recover the detailed structure such as the corner of the desk and the limb of the skeleton.

We further evaluated the generalization ability by testing the model on the scenes containing more objects than the ones in the training data. As shown in Fig. 6, the first two rows are the example of training data of the scenes containing two objects, and the last four rows demonstrate the generation results of the models on the scene containing four objects. We observe that GQN fails to generate the scene with four objects and tend to map the scene representations to the scenes containing two objects. In contrast, GRNN and the proposed STR-GQN successfully generalize the knowledge learned from the scene containing fewer objects. The reason might be that GRNN and the proposed STR-GQN store the object-specific information of different small regions in 3D space, while GQN directly learns the mapping from the distribution of latent codes to the variation of scenes containing two objects.

|       | GQN          | E-GQN        | STR-GQN      |
|-------|--------------|--------------|--------------|
| Mean Absolute Error (pixels) | | | |
| RRC   | 7.40±6.22    | **3.59±2.10** | 4.39±2.08   |
| SM7   | 3.13±1.30    | **2.14±0.53** | 3.11±0.93   |
| RFC   | 12.44±12.89  | 12.05±12.79  | **9.71±7.94** |
| RRO   | 10.12±5.15   | **6.59±3.23** | 7.17±3.24   |
| DISCO | 18.86±7.16   | **12.46±9.27** | 13.55±5.03 |
| Root Mean Squared Error (pixels) | | | |
| RRC   | 14.62±12.77  | **6.8±5.23** | 8.52±4.43   |
| SM7   | 9.97±4.34    | **5.63±2.21** | 10.56±3.02  |
| RFC   | 26.80±21.35  | 27.65±20.72  | **17.01±13.22** |
| RRO   | 19.63±9.14   | **12.08±6.52** | 13.78±5.87 |
| DISCO | 32.72±6.32   | **22.04±11.08** | 23.57±5.13 |

Table 1. Comparison of GQN, EGQN and STR-GQN in terms of the root mean square error.

|          | GQN          | GRNN         | STR-GQN         |
|----------|--------------|--------------|-----------------|
| ShapeNet | 0.109±0.029  | 0.084±0.017  | **0.079±0.016** |
| SM7      | 0.081±0.017  | 0.073±0.014  | **0.072±0.012** |
| RRC      | 0.506±0.046  | 0.497±0.047  | **0.494±0.019** |

Table 2. Comparison of GQN, GRNN and STR-GQN in terms of the cross entropy error based on the setting without stochastic unit.

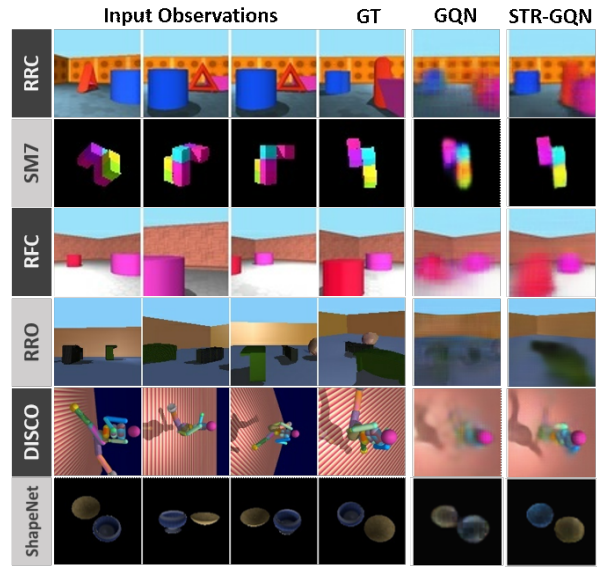

Figure 5. The rendering results of the proposed STR-GQN.

## 4.3. Scene Arithmetic

We conducted scene arithmetic experiments as introduced in [22] to prove that each world cell corresponds to an actual 3D location. In Fig. 7, column (A) shows the scenes containing two objects, column (B) shows the scenes containing one object appearing in the first column, and column (C) shows the scenes containing an additional object. After subtracting the scene representation of column (A) by the scene representation of column (B) and adding the scene representation of column (C), the results are expected to contain the remained object in column (A) and the object in column (C). Compared to GQN, GRNN and the proposed STR-GQN generates more reasonable results in which each object appears at the right location. However, the results of GRNN is less blurry than the proposed STR-GQN model. The reason might be that there are only 2048 world cells in our model, while the resolution of volumetric memory in GRNN is $32 \times 32 \times 32$ and can store finer details of the 3D space. Although the real 3D location of each world cell is untraceable, the experimental results prove that the proposed STR-GQN learns to store the information of different spatial regions in 3D space into different cells.

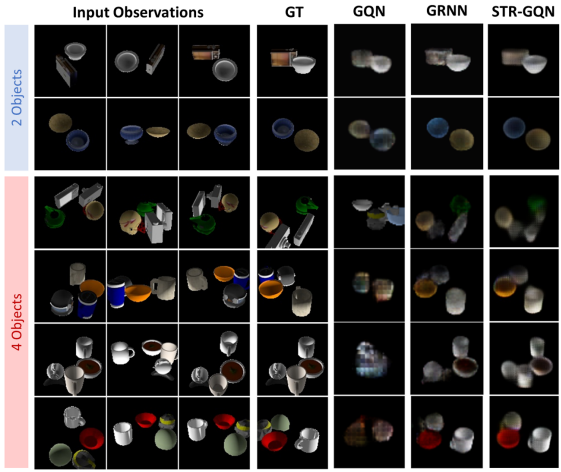

Figure 6. The rendering results of testing the model on the scenes containing more objects than the ones in the training data.

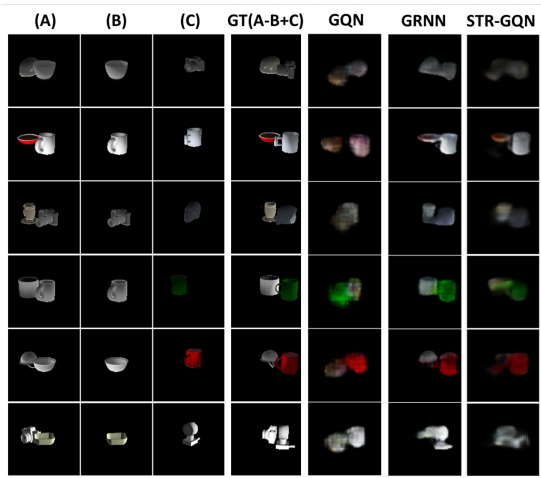

Figure 7. The rendering results of scene arithmetic.

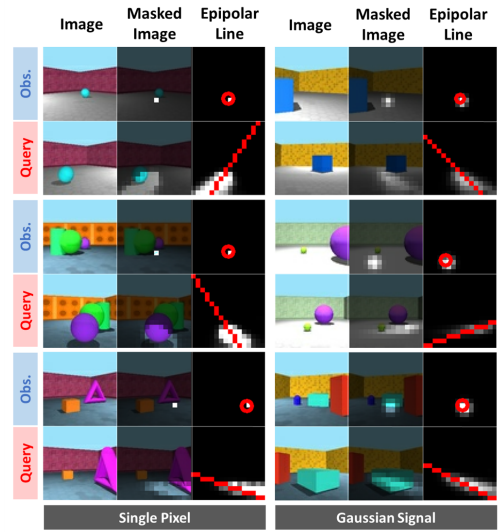

Figure 8. The visualization results of the routing process for one-single-pixel signal and Gaussian signals respectively.

### 4.4. Visualization of Routing Process

We demonstrate the visualization of the learned routing process for the proposed STRN. We sampled two poses of a scene as the input of STRN to generate the routing weights and then sent a signal through the routing process to observe how the information spreads onto the query view. Fig. 8 illustrate the visualization results for one-single-pixel signals and Gaussian signals respectively, in which the first column shows the images of the corresponding observation and query pose, the second column shows the images masked with the input signal for the observation and the spread signal for the query pose, and the third column shows the center position of the input signal and the corresponding epipolar line (computed by conventional geometric models) for the query pose (indicated by a red line).

We observe that the spread signals are consistent with the epipolar lines of the center positions of input signals, which means the proposed STRN successfully learns the spatial transformation in 3D space. In addition, the visualization results show that the signal of a pixel in the observation spreads onto an "area" rather than a line in the view space, which reveals the property of area-projection for a pixel. We also found that the STRN in view-to-world routing and the one in world-to-view routing should be the same (with symmetric relation). Otherwise, the proposed STR-GQN will converge slowly and the spread signals would be cluttered. To sum up, the visualization of routing process results proves that the spatial transformation can be learned purely from observation without applying prior knowledge of the perspective law and the calibrating parameters.

### 4.5. Different Imaging Situations

To evaluate the sensor adaptability of the model, we applied several image processing operations on the RRC dataset (as shown in Fig. 9) to simulate different imag-ing processes. We used the processed images to train/test the model and computed the pixel cross-entropy error as shown in Tab. 3. The increase of error compared to the model trained with the original dataset is also reported in the parenthesis. Compare to GRNN, the proposed STR-GQN has better performance in terms of pixel cross-entropy error and was less affected by the image processing operations.

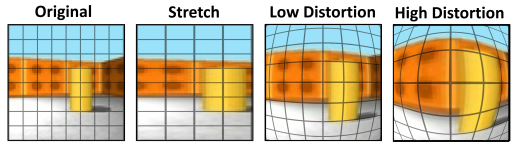

Figure 9. Examples of applying different image processing operations on RRC dataset.

|  | GQN | GRNN | STR-GQN |
|---|---|---|---|
| Original | 0.506 | 0.497 | 0.494 |
| Stretch | 0.510 (0.004) | 0.510 (0.013) | **0.504 (0.010)** |
| Low Dist. | 0.521 (0.015) | 0.515 (0.018) | **0.506 (0.012)** |
| High Dist. | 0.538 (0.032) | 0.537 (0.040) | **0.522 (0.028)** |

Table 3. The cross-entropy error of the models trained on RRC dataset with different image processing operations.

|  | Sum | Norm | OCM |
|---|---|---|---|
| 3 Obs. | 9.36 | 9.40 | 9.36 |
| 4 Obs. | 9.05 (-0.31) | 9.04 (-0.36) | 9.02 (-0.34) |
| 5 Obs. | 9.02 (-0.34) | 8.81 (-0.59) | 8.57 (-0.79) |
| 6 Obs. | 9.38 (+0.02) | 8.69 (-0.71) | 8.51 (-0.85) |
| 7 Obs. | 10.32 (+0.96) | 8.57 (-0.83) | 8.45 (-0.91) |
| 8 Obs. | 12.18 (+2.82) | 8.47 (-0.93) | 8.40 (-0.96) |

Table 4. The root mean squared error given different number of observation images for each fusion method based on RRC dataset.

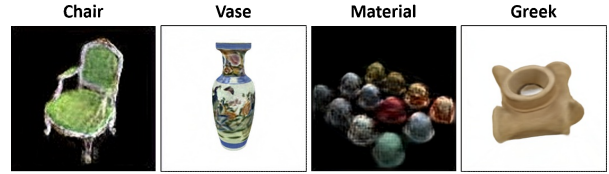

Figure 10. Some rendering results of the objects with complex texture or light condition.

### 4.6. Scene Fusion Operation

To compare different fusion methods, we investigated the inference results given a different numbers of observation images based on RRC dataset. Note that we randomly selected 1 to 5 observation images for training the model. In addition to the "Sum" fusion used in GQN and the proposed OCM, we also adopt the "Norm" fusion that applies the L2-normalization after summing the scene representations. Tab. 4 shows the root mean squared error of different fusion methods based on different number of observation images. The values reported in the parenthesis are the increase/decrease performance compared to the results based on 3 observation images. The performance of the "Sum" fusion method decreases when the number of observation images is larger than 5. The proposed OCM and "Norm" fusion methods can adapt to larger number of observation images than the one used in the training process. Note that we only trained the models with 0.8M steps in this experiment, thus the values in Tab. 4 are different from the ones in Tab. 1. Although the "Norm" fusion method achieves similar performance as the proposed OCM, OCM gives the scene representation a probabilistic explanation, which can be further applied to other tasks such as estimating the entropy difference for selecting the next best view.

### 4.7. View Synthesis for Complex Scenes

The datasets proposed in [3, 19, 22] are simple synthetic scenes without complex texture or light condition. To evaluate whether the proposed STR-GQN can be applied in more complex scenes, we trained the proposed model in a discriminative manner based on the "Vase" and "Greek" datasets proposed in [14] and the "Chair" and "Material" datasets proposed in [9]. Fig. 10 demonstrates some rendering results of the above datasets. Due to the page limit, the complete generated results and the corresponding analysis are shown in supplementary materials.

## 5. Discussion and Future Works

Based on the STR mechanism, the proposed STR-GQN successfully preserves the 3D rationality without applying any geometric priors and additional calibration parameters. The visualization results of the spread signal are consistent with the epipolar line, which demonstrates that the STRN learns the knowledge of transformation and projection. The scene arithmetic results prove that each world cell of the proposed model stores the object-specific information at a specific location in 3D space. This property is similar to the volumetric memory in GRNN. We look forward to replacing the geometry-aware module of spatially related tasks (such as visual odometry and 3d reconstruction) with the proposed STR mechanism. The proposed STR-GQN has the limitation of failing to recover the detailed structure and the texture whose appearance changes with view points. The main reason is that the proposed STR mechanism considers the features at a specific 3D location as the same for different viewpoints. A possible future direction is combining the proposed method with the capsule network [7] to model the equivariant features rather than the invariant features for the rotation so that we can utilize fewer cells to store more information of the scene.

# References

[1] Xu Chen, Jie Song, and Otmar Hilliges. Monocular neural image based rendering with continuous view control. In *Proceedings of the IEEE/CVF International Conference on Computer Vision*, pages 4090–4100, 2019.

[2] Inchang Choi, Orazio Gallo, Alejandro Troccoli, Min H Kim, and Jan Kautz. Extreme view synthesis. In *Proceedings of the IEEE/CVF International Conference on Computer Vision*, pages 7781–7790, 2019.

[3] SM Ali Eslami, Danilo Jimenez Rezende, Frederic Besse, Fabio Viola, Ari S Morcos, Marta Garnelo, Avraham Ruderman, Andrei A Rusu, Ivo Danihelka, Karol Gregor, et al. Neural scene representation and rendering. *Science*, 360(6394):1204–1210, 2018.

[4] Clément Godard, Oisin Mac Aodha, and Gabriel J Brostow. Unsupervised monocular depth estimation with left-right consistency. In *Proceedings of the IEEE Conference on Computer Vision and Pattern Recognition*, pages 270–279, 2017.

[5] Ariel Gordon, Hanhan Li, Rico Jonschkowski, and Anelia Angelova. Depth from videos in the wild: Unsupervised monocular depth learning from unknown cameras. In *Proceedings of the IEEE/CVF International Conference on Computer Vision*, pages 8977–8986, 2019.

[6] Karol Gregor, Frederic Besse, Danilo Jimenez Rezende, Ivo Danihelka, and Daan Wierstra. Towards conceptual compression. In *Advances In Neural Information Processing Systems*, pages 3549–3557, 2016.

[7] Geoffrey E Hinton, Sara Sabour, and Nicholas Frosst. Matrix capsules with em routing. In *International conference on learning representations*, 2018.

[8] Stephen Lombardi, Tomas Simon, Jason Saragih, Gabriel Schwartz, Andreas Lehrmann, and Yaser Sheikh. Neural volumes: Learning dynamic renderable volumes from images. *arXiv preprint arXiv:1906.07751*, 2019.

[9] Ben Mildenhall, Pratul P Srinivasan, Matthew Tancik, Jonathan T Barron, Ravi Ramamoorthi, and Ren Ng. Nerf: Representing scenes as neural radiance fields for view synthesis. *arXiv preprint arXiv:2003.08934*, 2020.

[10] Raul Mur-Artal, Jose Maria Martinez Montiel, and Juan D Tardos. Orb-slam: a versatile and accurate monocular slam system. *IEEE transactions on robotics*, 31(5):1147–1163, 2015.

[11] Kyle Olszewski, Sergey Tulyakov, Oliver Woodford, Hao Li, and Linjie Luo. Transformable bottleneck networks. In *Proceedings of the IEEE/CVF International Conference on Computer Vision*, pages 7648–7657, 2019.

[12] Eunbyung Park, Jimei Yang, Ersin Yumer, Duygu Ceylan, and Alexander C Berg. Transformation-grounded image generation network for novel 3d view synthesis. In *Proceedings of the ieee conference on computer vision and pattern recognition*, pages 3500–3509, 2017.

[13] Johannes L Schonberger and Jan-Michael Frahm. Structure-from-motion revisited. In *Proceedings of the IEEE conference on computer vision and pattern recognition*, pages 4104–4113, 2016.

[14] Vincent Sitzmann, Justus Thies, Felix Heide, Matthias Nießner, Gordon Wetzstein, and Michael Zollhofer. Deepvoxels: Learning persistent 3d feature embeddings. In *Proceedings of the IEEE Conference on Computer Vision and Pattern Recognition*, pages 2437–2446, 2019.

[15] Vincent Sitzmann, Michael Zollhöfer, and Gordon Wetzstein. Scene representation networks: Continuous 3d-structure-aware neural scene representations. In *Advances in Neural Information Processing Systems*, pages 1121–1132, 2019.

[16] Shao-Hua Sun, Minyoung Huh, Yuan-Hong Liao, Ning Zhang, and Joseph J Lim. Multi-view to novel view: Synthesizing novel views with self-learned confidence. In *Proceedings of the European Conference on Computer Vision (ECCV)*, pages 155–171, 2018.

[17] Maxim Tatarchenko, Alexey Dosovitskiy, and Thomas Brox. Multi-view 3d models from single images with a convolutional network. In *European Conference on Computer Vision*, pages 322–337. Springer, 2016.

[18] Sebastian Thrun. Probabilistic robotics. *Communications of the ACM*, 45(3):52–57, 2002.

[19] Joshua Tobin, Wojciech Zaremba, and Pieter Abbeel. Geometry-aware neural rendering. In *Advances in Neural Information Processing Systems*, pages 11559–11569, 2019.

[20] Shubham Tulsiani, Alexei A Efros, and Jitendra Malik. Multi-view consistency as supervisory signal for learning shape and pose prediction. In *Proceedings of the IEEE conference on computer vision and pattern recognition*, pages 2897–2905, 2018.

[21] Shubham Tulsiani, Tinghui Zhou, Alexei A Efros, and Jitendra Malik. Multi-view supervision for single-view reconstruction via differentiable ray consistency. In *Proceedings of the IEEE conference on computer vision and pattern recognition*, pages 2626–2634, 2017.

[22] Hsiao-Yu Fish Tung, Ricson Cheng, and Katerina Fragkiadaki. Learning spatial common sense with geometry-aware recurrent networks. In *Proceedings of the IEEE Conference on Computer Vision and Pattern Recognition*, pages 2595–2603, 2019.

[23] Olivia Wiles, Georgia Gkioxari, Richard Szeliski, and Justin Johnson. Synsin: End-to-end view synthesis from a single image. In *Proceedings of the IEEE/CVF Conference on Computer Vision and Pattern Recognition*, pages 7467–7477, 2020.

[24] Jiajun Wu, Yifan Wang, Tianfan Xue, Xingyuan Sun, William T Freeman, and Joshua B Tenenbaum. Marrnet: 3d shape reconstruction via 2.5 d sketches. *arXiv preprint arXiv:1711.03129*, 2017.

[25] Xinchen Yan, Jimei Yang, Ersin Yumer, Yijie Guo, and Honglak Lee. Perspective transformer nets: Learning single-view 3d object reconstruction without 3d supervision. In *Advances in neural information processing systems*, pages 1696–1704, 2016.

[26] Huangying Zhan, Ravi Garg, Chamara Saroj Weerasekera, Kejie Li, Harsh Agarwal, and Ian Reid. Unsupervised learning of monocular depth estimation and visual odometry with deep feature reconstruction. In *Proceedings of the IEEE*

*Conference on Computer Vision and Pattern Recognition*, pages 340–349, 2018.

[27] Tinghui Zhou, Matthew Brown, Noah Snavely, and David G Lowe. Unsupervised learning of depth and ego-motion from video. In *Proceedings of the IEEE Conference on Computer Vision and Pattern Recognition*, pages 1851–1858, 2017.

[28] Tinghui Zhou, Shubham Tulsiani, Weilun Sun, Jitendra Malik, and Alexei A Efros. View synthesis by appearance flow. In *European conference on computer vision*, pages 286–301. Springer, 2016.

# STR-GQN: Scene Representation and Rendering for Unknown Cameras Based on Spatial Transformation Routing
## – Supplementary Materials –

## 1. Occupancy Concept Mapping

The proof of the proposed OCM (cf. Sec. 3.3) and the analysis of routing process in probabilistic perspective are given as follows. Let $o_{k,c}$ denote the random variable indicating a concept exists at the 3D location of the $k$th world cell (i.e. $o_{k,c} = 1$) or not (i.e. $o_{k,c} = 0$), and $o_{ij,c}$ denote the random variable indicating a concept $c$ exists at the 2D location $(i, j)$ of a view cell. The goal of OCM is to estimate the scene cell $sc_{k,c}$, which represents the posterior of the existing probability for a concept $c$ at the location of the $k$th world cell given the observed images $x^{1:N}$:

$$sc_{k,c} = p(o_{k,c} = 1 | x^{1:N}). \tag{1}$$

The log-odds of the concept existing probability in world space is defined by

$$\log Odd(o_{k,c}) = \log \frac{p(o_{k,c} = 1)}{p(o_{k,c} = 0)}. \tag{2}$$

The posterior of the log-odds given an observation can be computed via Bayes' theorem:

$$\log Odd(o_{k,c}|x) = \log \frac{p(x|o_{k,c}=1)p(o_{k,c}=1)/p(x)}{p(x|o_{k,c}=0)p(o_{k,c}=0)/p(x)}$$
$$= \log \frac{p(x|o_{k,c}=1)}{p(x|o_{k,c}=0)} Odd(o_{k,c})$$
$$= \log \frac{p(x|o_{k,c}=1)}{p(x|o_{k,c}=0)} + \log Odd(o_{k,c}). \tag{3}$$

Without any observation, the initial prior of the log-odds for the concept existing probability is zero:

$$\log Odd(o_{k,c}) = \log \frac{p(o_{k,c}=1)}{p(o_{k,c}=0)} = \log \frac{0.5}{0.5} = 0. \tag{4}$$

Let $wc_{k,c}$ denote the value of the $c$th channel for the $k$th world cell, which is assumed as the log of likelihood ratio of a concept:

$$wc_{k,c} = \log \frac{p(x|o_{k,c}=1)}{p(x|o_{k,c}=0)}. \tag{5}$$

The posterior of log-odds given $N$ observations $x^1, x^2, ..., x^N$ can be estimated via iterative Bayesian updating according to Eq. 3 and 4:

$$\log Odd(o_{k,c}|x^1, x^2, ..., x^N)$$
$$= (...((\log Odd(o_{k,c}) + wc_{k,c}^1) + wc_{k,c}^2) + ...) + wc_{k,c}^N$$
$$= \sum_{n=1}^{N} wc_{k,c}^n, \tag{6}$$

where $n$ represents the index of different observations. The scene cell is then calculated by the posterior of log-odds:

$$\log Odd(o_{k,c}|x^{1:N}) = \log \frac{p(o_{k,c}=1|x^{1:N})}{1 - p(o_{k,c}=1|x^{1:N})},$$
$$sc_{k,c} = p(o_{k,c}=1|x^{1:N})$$
$$= \frac{\exp(\log Odd(o_{k,c}|x^{1:N}))}{1 + \exp(\log Odd(o_{k,c}|x^{1:N}))}$$
$$= \sigma(\log Odd(o_{k,c}|x^{1:N})) = \sigma(\sum_{n=1}^{N} wc_{k,c}^n). \tag{7}$$

We further analyze the meanings of the routing process under the mathematical framework of OCM. Similar to $wc_{k,c}$, let $vc_{ij,c}$ denote the value of the $c$th channel for a view cell at the 2d location $(i, j)$, which is assumed as the log of likelihood ratio of concept $c$:

$$vc_{ij,c} = \log \frac{p(x|o_{ij,c}=1)}{p(x|o_{ij,c}=0)}. \tag{8}$$

The routing process transforms the message from the view cells to the world cells. The information passing to the $k$th world cell is defined by the weighted sum of each view cell based on the weighting $p_k^{dist}(i,j)$:

$$\begin{aligned}
wc_{k,c} &= \sum_{i,j} p_k^{dist}(i,j) vc_{ij,c} \\
&= \sum_{i,j} p_k^{dist}(i,j) \log \frac{p(x|o_{ij,c}=1)}{p(x|o_{ij,c}=0)} \\
&= \log \prod_{i,j} \left( \frac{p(x|o_{ij,c}=1)}{p(x|o_{ij,c}=0)} \right)^{p_k^{dist}(i,j)} \\
&= \log \prod_{i,j} \left( \frac{p(o_{ij,c}=1|x)p(x)/p(o_{ij,c}=1)}{p(o_{ij,c}=0|x)p(x)/p(o_{ij,c}=0)} \right)^{p_k^{dist}(i,j)}
\end{aligned} \quad (9)$$

Moreover, the Eq. 5 can be re-written as:

$$wc_{k,c} = \log \frac{p(o_{k,c}=1|x)p(x)/p(o_{k,c}=1)}{p(o_{k,c}=0|x)p(x)/p(o_{k,c}=0)}. \quad (10)$$

Combining Eq. 9 and 10, we observe that the probability distribution $p_k^{dist}(i,j)$ can be the weighting of the weighted geometry mean for the concept existing probability $p(o_{ij,c})$ in view space:

$$\begin{aligned}
p(o_{k,c}) &= \sqrt[\sum_{i,j} p_k^{dist}(i,j)]{\prod_{i,j} p(o_{ij,c})^{p_k^{dist}(i,j)}} \\
&= \prod_{i,j} p(o_{ij,c})^{p_k^{dist}(i,j)}.
\end{aligned} \quad (11)$$

## 2. Additional Rendering Results

Fig. 2 to Fig. 6 demonstrate more rendering results of the experiments introduced in Sec. 4.2 for each dataset.

## 3. Visualization of Routing Process

Fig. 7 demonstrates more routing results of the experiments introduced in Sec. 4.4. Furthermore, to evaluate whether the proposed STRN can learn the general mapping between world cells and the continuous view space (as mentioned in Sec. 3.2), we trained the model with $16 \times 16$ view cells and visualized the routing results of view cells for different resolutions by sampling and interpolating the 2D location codes. As Fig. 8 shows, the proposed STRN can learn the routing weights for different resolution of view cells.

## 4. Results for Complex Scenes

As mentioned in Sec. 4.7, to evaluate whether the proposed STR-GQN can be applied in more complex scenes, we trained the proposed model in a discriminative manner based on the "Vase", "Greek", "Chair", and "Material" datasets. We randomly sampled 32 to 48 frames as the observation images for training and always utilize 48 frames as the observation images for testing. Fig. 9 to 12 demonstrate the experimental results for each dataset.

The proposed STR-GQN achieves good performance on the "chair" dataset and "vase" dataset. The results show that the proposed model can reconstruct the complex texture when the 3D structures are simple. However, the proposed model fails to generate clear results for "material" and "greek" datasets, which reveals the limitation of our model. The texture of "material" dataset is simpler than "vase" dataset while it changes with the view pose. The proposed model fails to reconstruct this kind of texture because the STR mechanism only considers the view-independent concepts in 3D space. On the other hand, the texture of "greek" dataset is simple and view-independent, but its 3D structure is complex. The proposed model fails to reconstruct the detailed structure due to the limited number of world cells.

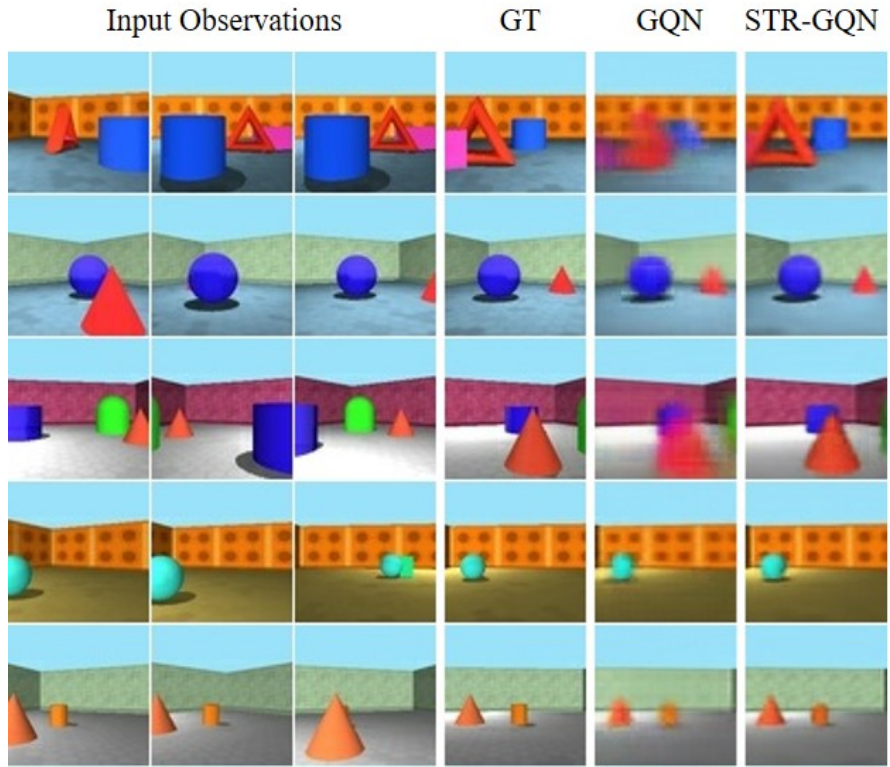

Figure 1. Rendering results of RRC dataset.

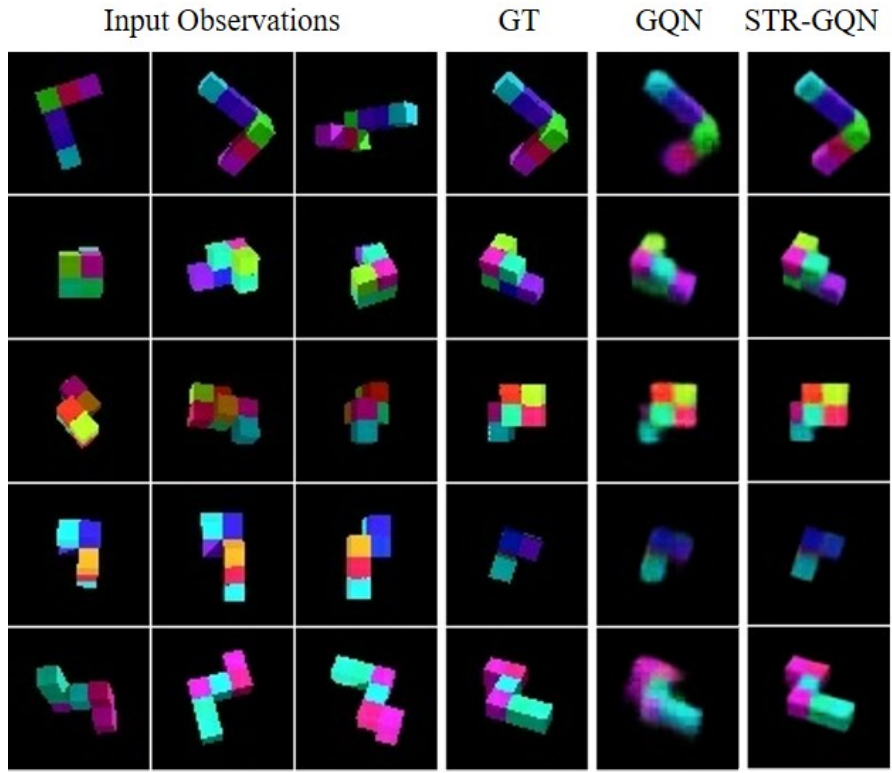

Figure 2. Rendering results of SM7 dataset.

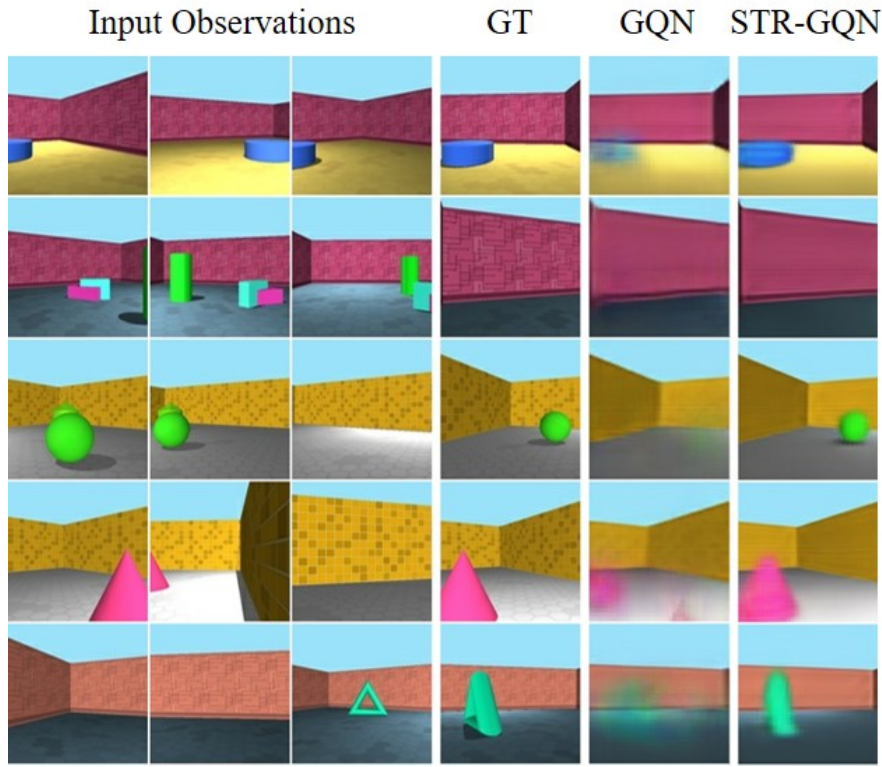
Figure 3. Rendering results of RFC dataset.

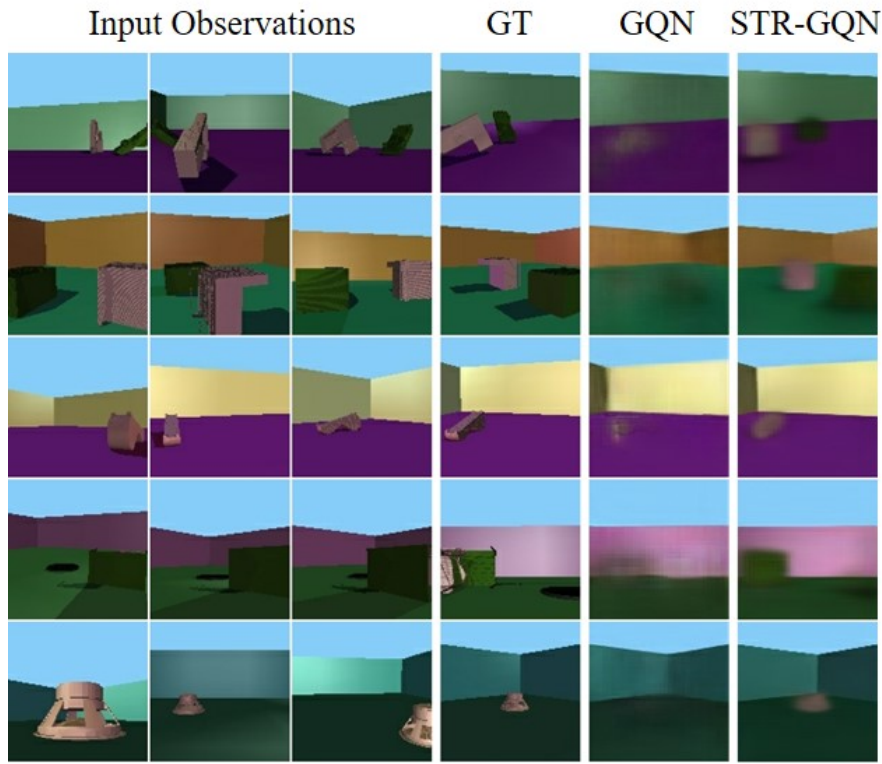
Figure 4. Rendering results of RRO dataset.

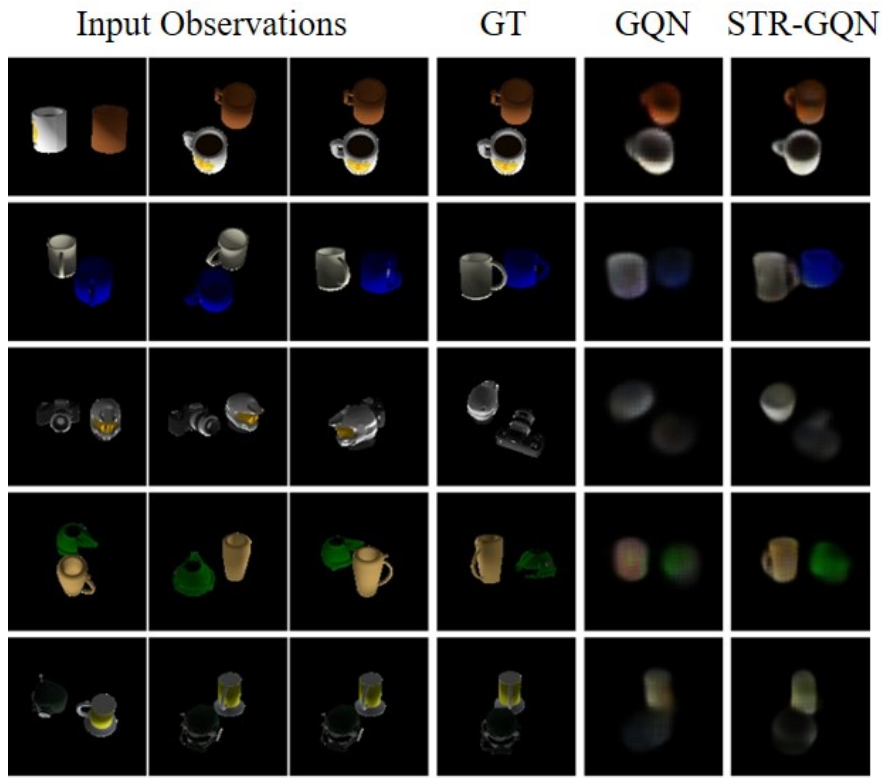

Figure 5. Rendering results of ShapeNet dataset containing 2 objects.

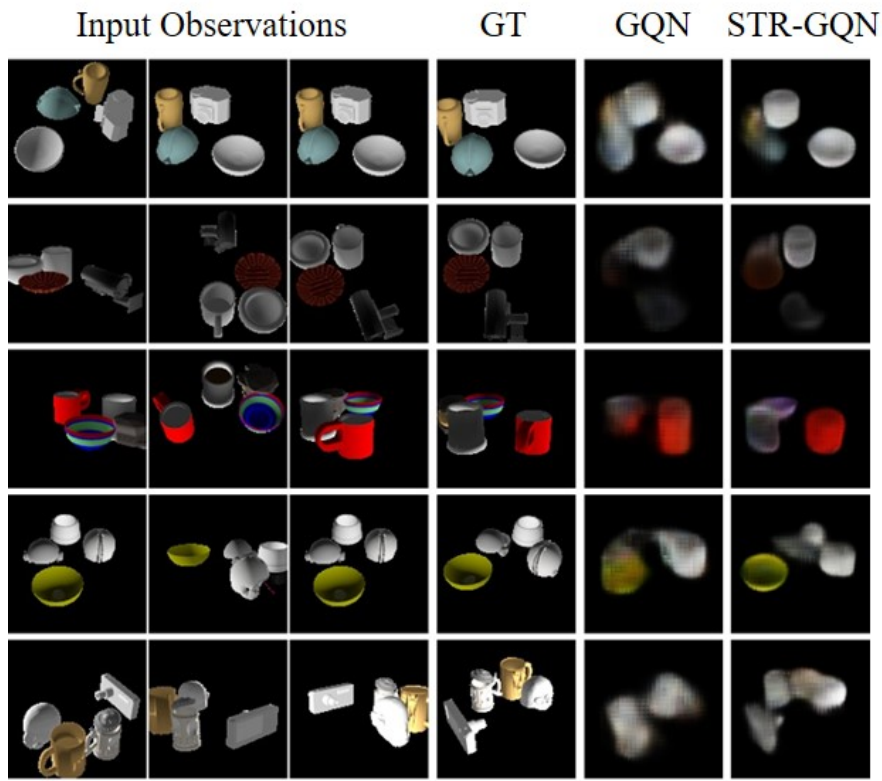

Figure 6. Rendering results of ShapeNet dataset containing 4 objects.

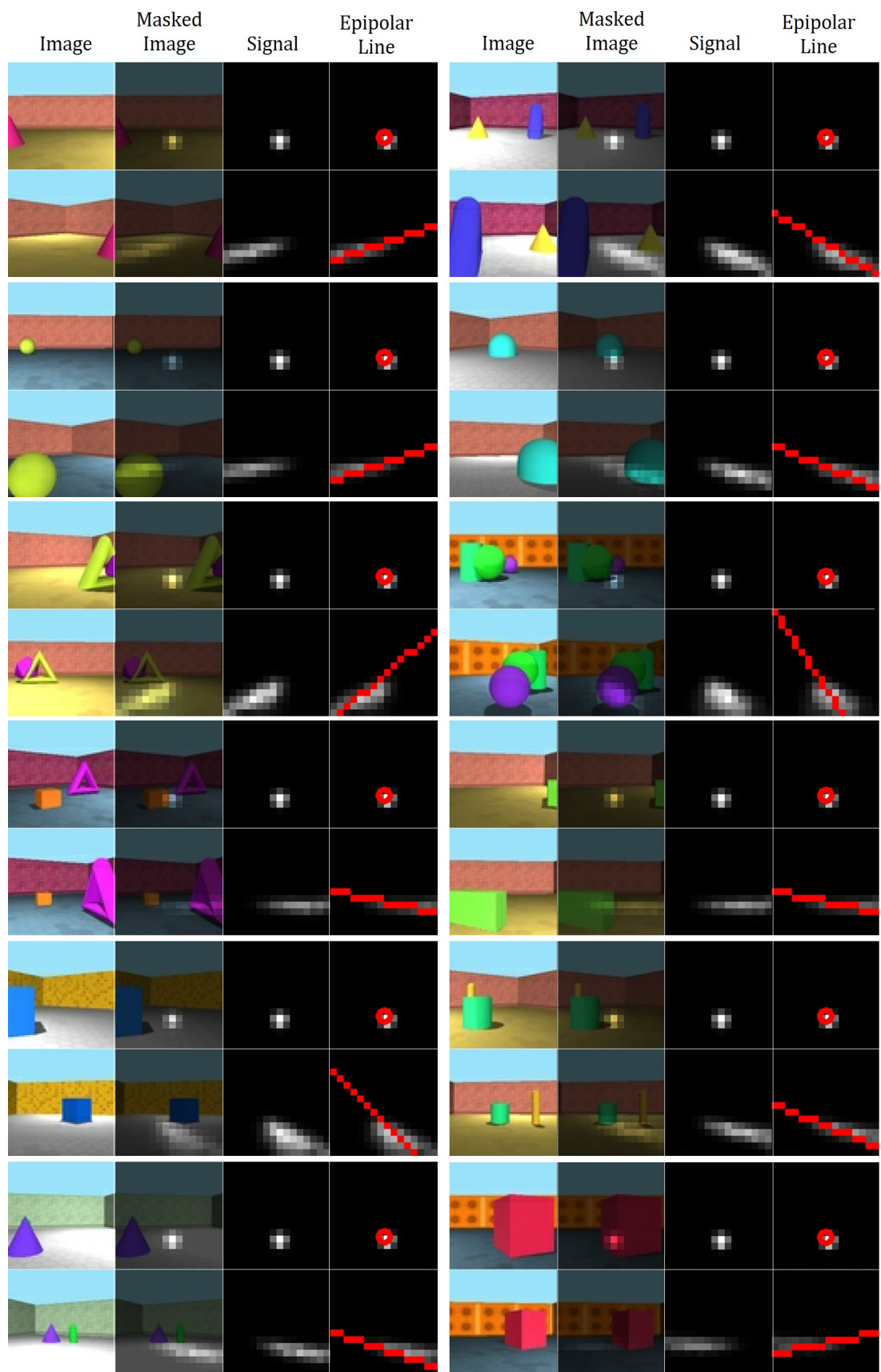

Figure 7. Visualization of the routing process.

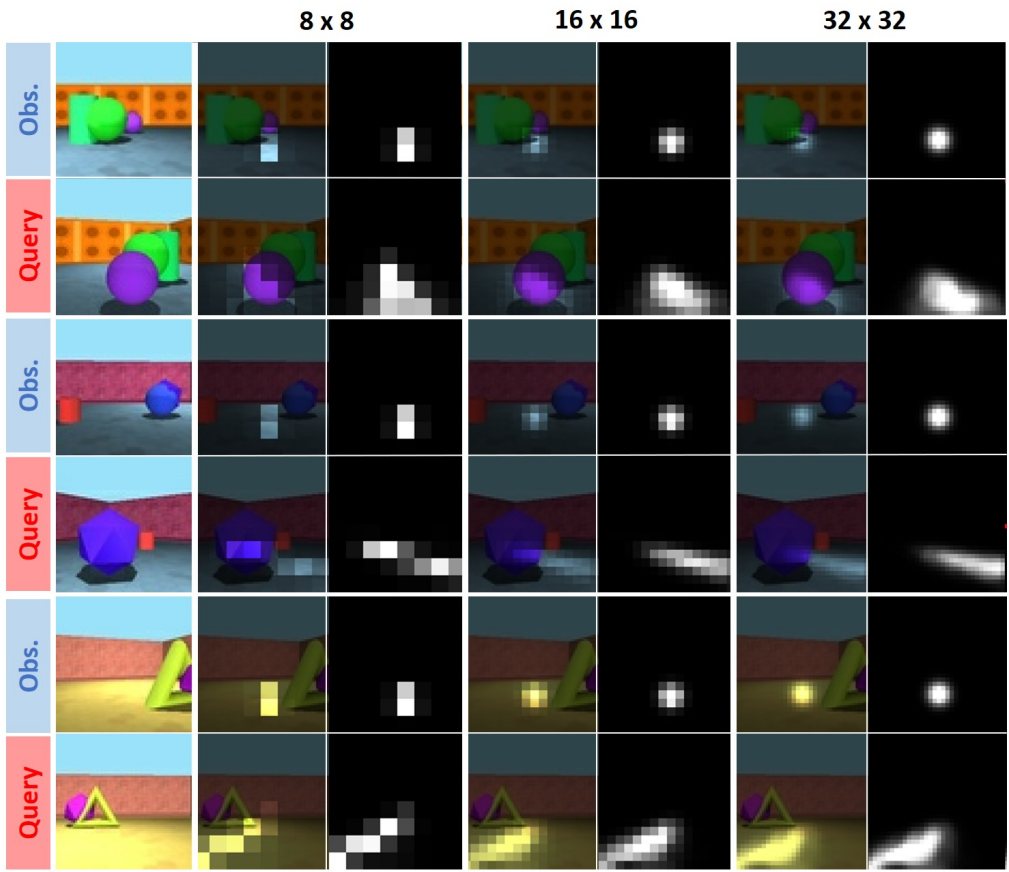

Figure 8. Visualization of the routing process for different resolution of view space.

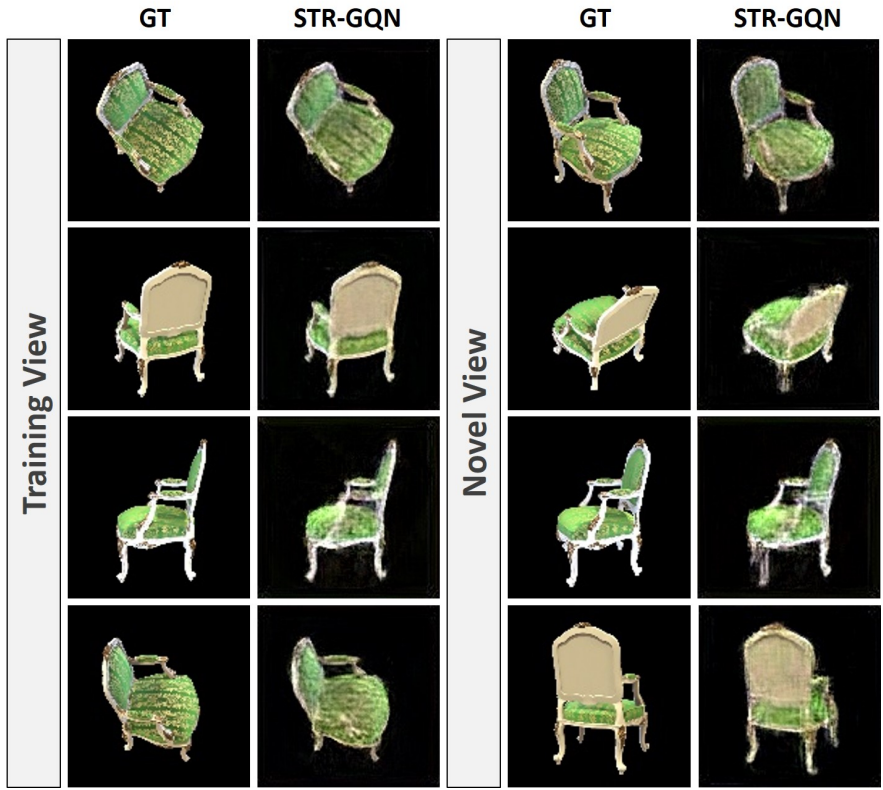

Figure 9. Rendering results of "Chair" dataset.

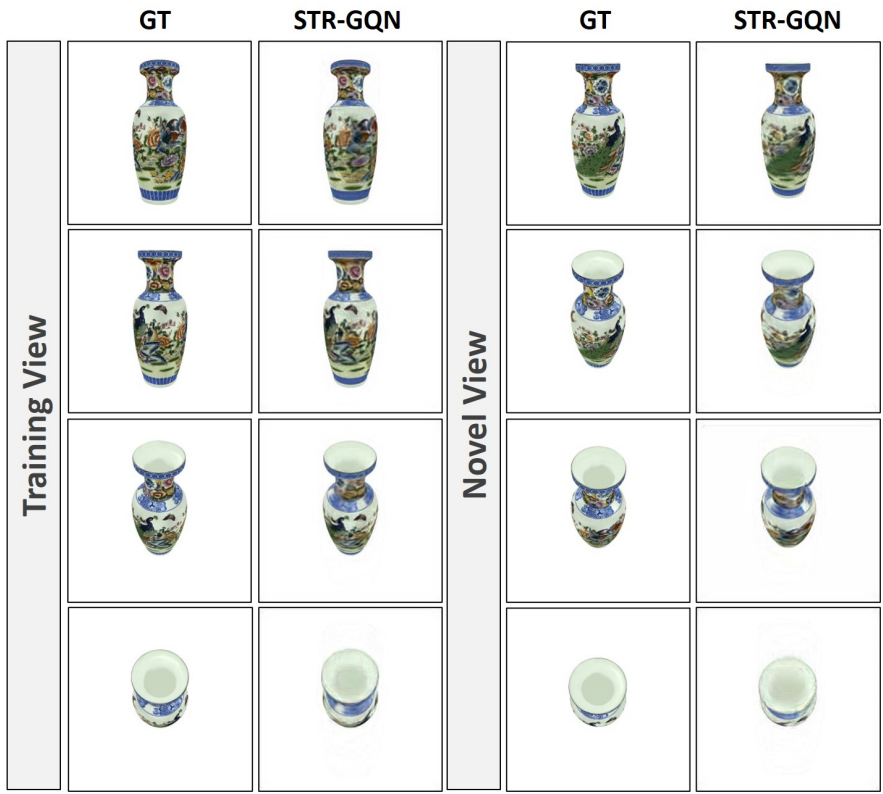

Figure 10. Rendering results of "Vase" dataset.

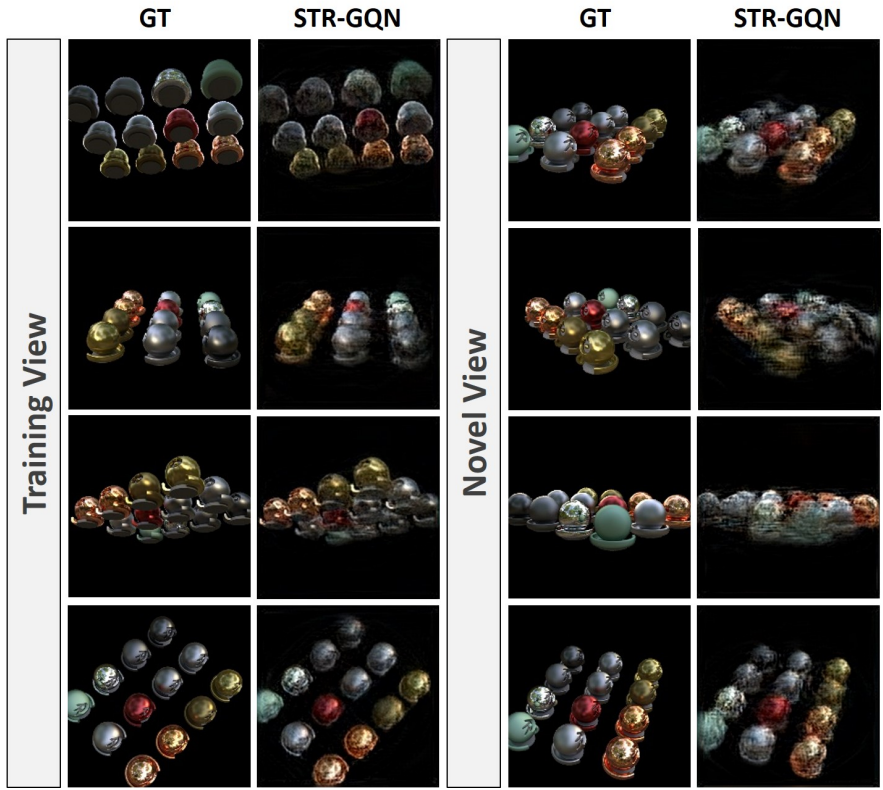

Figure 11. Rendering results of "Material" dataset.

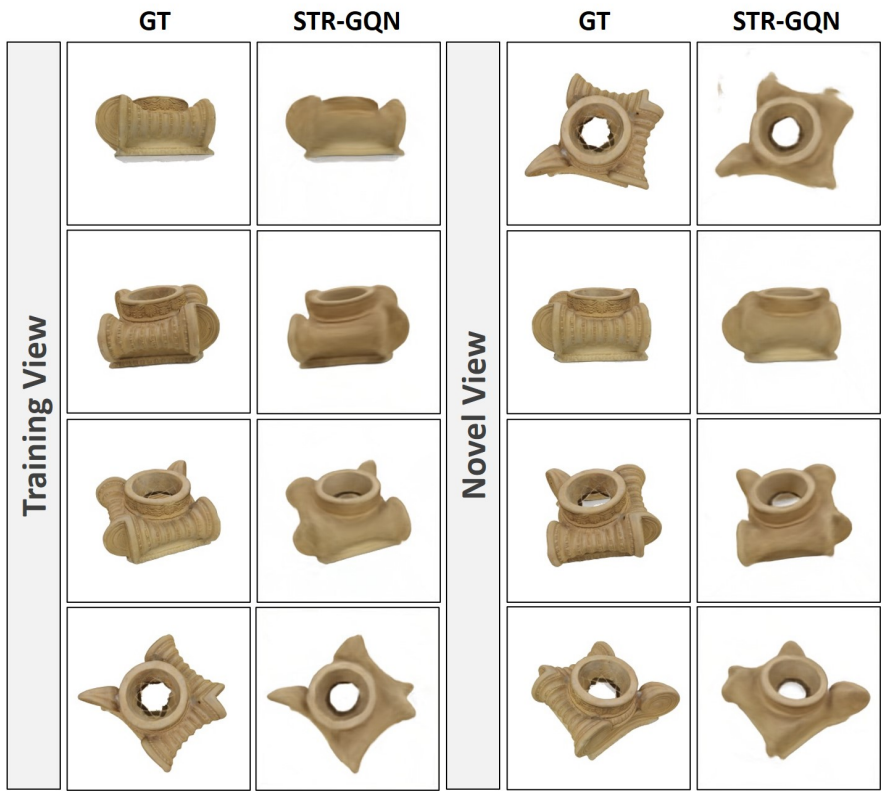

Figure 12. Rendering results of "Greek" dataset.